\documentclass[letterpaper]{article} 
\usepackage{aaai23}  
\usepackage{times}  
\usepackage{helvet}  
\usepackage{courier}  
\usepackage[hyphens]{url}  
\usepackage{graphicx} 
\usepackage{natbib}  
\usepackage{caption} 
\usepackage{algorithm}
\usepackage{algorithmic}

\usepackage{amssymb}
\usepackage{amsmath}
\usepackage{multirow}
\newcommand{\paren}[1]{\left( #1 \right)}

\newcommand{\brackets}[1]{\left[ #1 \right]}

\usepackage{newfloat}
\usepackage{listings}
\DeclareCaptionStyle{ruled}{labelfont=normalfont,labelsep=colon,strut=off} 
\floatstyle{ruled}
\newfloat{listing}{tb}{lst}{}
\floatname{listing}{Listing}
\title{DeepRHP: A Hybrid Variational Autoencoder for Designing Random Heteropolymers as Protein Mimics}

\author{
    Shuni Li, Zhiyuan Ruan, Andy Shen, Ivan Jayapurna, Ting Xu, Haiyan Huang
}
\affiliations{
    University of California Berkeley; Berkeley, California, 94720, USA.\\

    \{shuni\_li, ruanzy, aashen, ivanfj, tingxu, hyh0110\}@berkeley.edu
}

\nocopyright

\begin{document}

\maketitle

\begin{abstract}

Synthetic random heteropolymers (RHPs), consisting of a predefined set of monomers, offer an approach toward the design of protein-like materials. These RHPs, if designed appropriately, can mimic protein behavior and function. As such, there is a need for computational tools to efficiently guide RHP design. We bridge this gap by developing DeepRHP, a modified variational autoencoder (VAE) model under a semi-supervised framework. By equipping a classical VAE with an additional feature-based VAE, DeepRHP forces the latent space to capture structures of critical chemical features as well as individual RHP sequence patterns. In this sense, our method is versatile by allowing any relevant features to be incorporated in a hybrid manner. We demonstrate the effectiveness of DeepRHP by suggesting potential monomer compositions that stabilize membrane proteins (e.g. Aquaporin Z) in non-native environments and cross-validating our prediction with published results. The concordance between our model and true RHP function suggests strong potential in utilizing hybrid autoencoder architectures to guide RHP design for proteins and other biological compounds.

\end{abstract}

\section{Introduction}

There is a significant interest in engineering synthetic materials capable of replicating protein functions while satisfying stability and compatibility with device fabrication and integration. However, it remains an insurmountable challenge to synthesize sequence-specific polymers. This has led to a recent surge of research in designing protein-like random heteropolymers. Random heteropolymers (RHPs) are an ensemble of many polymer chains with each being composed of monomers arranged in random order \cite{Hilburg2020}. Recent developments have demonstrated that RHPs can act as chaperone proteins for protein stabilization in non-biological environments \cite{Panganiban2018}, a critical bottleneck to fabricate protein-embedded plastics for end-of-life plastic degradation \cite{DelRe2021}. In addition, RHPs can be designed to act as channel proteins for rapid and selective proton transportation \cite{Jiang2020}, important for fuel cells and energy storage. 

Despite the fact that RHPs can serve as great biofunctional materials, designing RHPs with desired function is challenging because both the exact monomeric sequences and conformations of synthetic RHP chains are not deterministic. Traditional protein design methods rely heavily on high-throughput sequencing data and 3D structures. For example, directed evolution methods evolve protein function by iteratively mutating a selected protein sequence \cite{Arnold2018}, while \textit{de novo} methods build novel proteins that fold into a certain structure \cite{Huang2016}. Without exact sequences and structures, there are no rational design principles for creating suitably functional RHP chains. Current RHP designs are largely empirical and depend on time-intensive lab screenings over various monomer compositions and chain lengths. For each RHP made in the lab, ensembles of thousands of sequences are simulated under the same monomer composition in order to understand why certain compositions perform better than others. In this process, scientists face two practical design 
questions that can potentially accelerate progress if answered:
\begin{itemize}
    \item How many monomers should be included in a RHP system? Recent results show that RHPs can mimic protein function with only four monomers \cite{Panganiban2018, Jiang2020}, but it remains unclear how many monomers are enough to include in the alphabet.
    \item How can one find monomer compositions corresponding to specific protein functions?
\end{itemize}

Answering these questions requires new methods to model and analyze RHP sequences as an ensemble instead of as individual chains. To our knowledge, there is very limited literature on computational methods of modeling RHPs. As the only two examples, \citeauthor{Zhou2022} \shortcite{Zhou2022} used Hidden Markov Models to characterize the functionality of proton-transporting RHPs and \citeauthor{Tamasi2022} \shortcite{Tamasi2022} utilized Gaussian process regression coupled with Bayesian optimization for optimal copolymer identification. 

Here we propose DeepRHP, a modified variational autoencoder trained in a semi-supervised manner, for modeling general RHP sequence data and discovering RHP compositions for protein function. This tool serves as a first step that can guide RHP design by examining their protein-mimicking behavior. The key contributions of this study are:

\begin{itemize}
\item We are the first to answer RHP design questions with deep learning. DeepRHP learns interpretable latent representations for RHP sequences and provides a platform to perform similarity analysis between target proteins and RHP sequences in an ensemble. 
\item DeepRHP provides insights into the two important design parameters: monomer alphabet size and monomer composition. We show that the best monomer composition suggested by DeepRHP matches published experimental results.
\item DeepRHP is flexible enough to incorporate any function-related chemical features for a wide variety of protein functions.
\end{itemize}

VAE-based architectures are some of the first model classes used to identify latent representations for biological sequences, and are useful in downstream tasks like identifying mutation effects \cite{Sinai2017, Riesselman2018} and designing novel functional proteins \cite{Greener2018, Costello2019}. Therefore, we should expect to leverage the same machine learning theory in macromolecular cheminformatics, specifically in this instance of using RHPs to mimic natural biopolymers.

\begin{figure*}[t]
    \centering
            \includegraphics[width=0.9\textwidth]{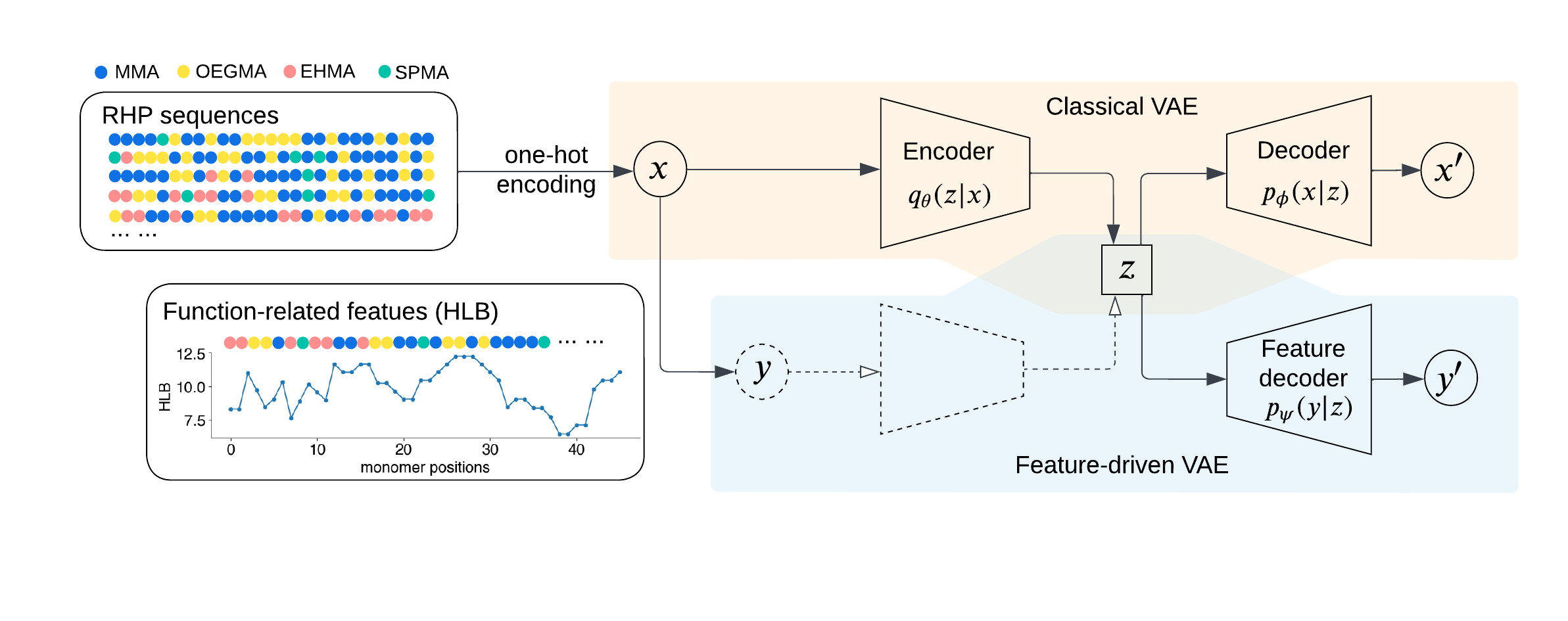}
    \caption{DeepRHP model architecture consisting of a classical VAE equipped with an additional feature-based VAE.}
    \label{fig:fig1}
\end{figure*}

\section{Data}

Our work utilizes the RHP system developed in both \citeauthor{Panganiban2018} \shortcite{Panganiban2018} and \citeauthor{Jiang2020} \shortcite{Jiang2020}. This system consists of four methacrylate-based monomers: methyl methacrylate (MMA), 2-ethylhexyl methacrylate (EHMA), oligo (ethylene glycol) methacrylate (OEGMA), and 3-sulfopropyl methacrylate potassium salt (SPMA). MMA and EHMA are the hydrophobic monomers used to tailor overall hydrophobicity, while OEGMA and SPMA are the hydrophilic monomers used to reduce the aggregation propensity of RHPs.

We used Compositional Drift, a software developed by \citeauthor{Smith2019} \shortcite{Smith2019} to simulate 10,000 sequences per monomer composition listed in Table \ref{table_comp}. This software uses established mathematical copolymer models in tandem with Monte-Carlo simulation to calculate RHP sequences based on experimental conditions. The authors showed that, while each chain simulated is random at the sequence level, it contains characteristic segments that have a well-defined statistical distribution \cite{Smith2019}. The reasoning behind the monomeric compositions for each specific RHP is further discussed in Section 4. 

We also collected 30,000 membrane protein sequences and 30,000 globular protein sequences with 50\% identity threshold from the UniProt database \cite{uniprot}. Some common pre-processing procedures were performed, including discarding sequences with uncommon amino acids and lengths. Each protein was then reduced into its monomer-equivalent form according to the assignment in Table \ref{table_aa2m}. Note that the reduction of protein alphabet is not uncommon in protein sequence analysis, see \citeauthor{Liang2022} \shortcite{Liang2022} for a comprehensive review. Here our reduction rule is based on monomer hydrophobicity and charge.

\begin{table}[t]
\centering
\begin{tabular}{l|l|l|l|l|l}
      & RHP & MMA & OEGMA & EHMA & SPMA \\
    \hline
    \multirow{5}{*}[-5.8pt]{2 Mon.}  & A & 0 & 10 & 90 & 0 \\
    & B & 0 & 30 & 70 & 0 \\
    & C & 0 & 50 & 50 & 0 \\
    & D & 0 & 70 & 30 & 0 \\
    & E & 0 & 90 & 10 & 0 \\
    \hline
    \multirow{5}{*}[-5.8pt]{4 Mon.}& 1 & 70 & 25 & 0 & 5 \\
    & 2 & 65 & 25 & 5 & 5 \\
    & 3 & 60 & 25 & 10 & 5 \\
    & 4 & 50 & 25 & 20 & 5 \\
    & 5 & 40 & 25 & 30 & 5 \\
    & 6 & 20 & 25 & 50 & 5 \\
    & 7 & 0 & 25 & 70 & 5 \\
\end{tabular}
\caption{Two and four-monomer composition of RHPs used for training}
\label{table_comp}
\end{table}

\begin{table}[t]
\centering
\begin{tabular}{l|l|l}
    Amino acid & Monomer equiv. & Property \\
    \hline
    C, Y, A, T, G & MMA & Hydrophobic \\
    S, Q, H, N, P & OEGMA & Hydrophilic \\
    L, I, F, W, V, M & EHMA & Very Hydrophobic \\
    E, D, R, K & SPMA & Charged \\
\end{tabular}
\caption{Amino acid (protein) to monomer (RHP) conversion}
\label{table_aa2m}
\end{table}

\section{DeepRHP Methodology}

In order to address the domain questions raised in Section 1, we developed DeepRHP, a modified variational autoencoder under semi-supervised framework for learning low-dimensional RHP sequence representations. The model architecture is illustrated in Figure \ref{fig:fig1}. We assume the sequence family $X$ follows a probability distribution $p(x)$ and there exists an underlying latent variable $z \sim N(\mu_z,  \Sigma_z)$ that captures intrinsic unobserved sequence properties. For each sequence $x$, there also exists a function-related feature $y$, which can be considered as a deterministic transformation of $x$. In the application case presented in Section 4, $y$ is the average hydrophilic–lipophilic balance (HLB) value of sliding windows along each sequence \cite{Kyte1982}. HLB measures local hydrophobicity and solubility distributions and is closely related to RHP functions \cite{Panganiban2018, Jiang2020}. The motivation for introducing other function-related chemical features (e.g. HLB) is for them to guide the formation of the latent space.

To incorporate a chemical feature $y$ into our VAE model, we add a feature-driven VAE in parallel with the classical VAE. $y$ and $x$ share the common latent variable $z$. This is equivalent to simultaneously training two VAEs with shared latent embeddings, and the encoder relies only on $x$ since $y$ is a direct transformation of $x$, as indicated by the dashed lines in Figure \ref{fig:fig1}.

The objective is still to maximize the log-likelihood $\log p(x)$ given sequence data $X$ as shown in equation:
\begin{equation}
    \label{eq:like1}
    \log p(x) = \log \int  p(x\mid z)~p(z) ~dz .
\end{equation}
Under the regular VAE setting, Equation \ref{eq:like1} can be bound by the well-known evidence lower bound (ELBO) \cite{kingma_auto-encoding_2014, rezende_stochastic_2014}:
\begin{equation}
  \label{eqn:elbo}
  \log p(x) \geq \mathbb{E}_{q}\brackets{\log p(x\mid z)} - D_{KL}\paren{q(z\mid x)~ \|~ p(z)},
\end{equation}
where $q$ is the learned posterior of the normal distribution family. In practice, $p$ and $q$ are learned by the encoder and decoder and their weights are optimized through gradient descent.

Traditionally, the reconstruction loss term is approximated by mean-squared error for continuous input, or cross-entropy loss for discrete input. By imposing this hybrid architecture, we can approximate the reconstruction loss through both the classical VAE on $x$, the feature-driven VAE on $y$, or a weighted sum of both. Our modified ELBO  that considers both sequence structures and chemical features is then formulated as
\begin{align}
  \label{eqn:loss}
  \log p(x) \geq \  & \alpha \mathbb{E}_{q}\brackets{\log p(x\mid z)}+ (1-\alpha) \mathbb{E}_{q}\brackets{\log p(y\mid z)} - \nonumber\\
  & D_{KL}\paren{q(z\mid x)~ \|~ p(z)},
\end{align}
where $\alpha$ is a hyperparameter that dictates how much weight is placed on each approximation term. In our case, the first two terms of Equation \ref{eqn:loss} are approximated as follows:
\begin{equation}
\mathbb{E}_{q}\brackets{\log p(x\mid z)} \approx \sum_x \sum_l p(x_l) * \log(p(x_l\mid z)
\end{equation}
\begin{equation}
\mathbb{E}_{q}\brackets{\log p(y\mid z)} ) \approx -\sum_y ||y - y'||_2^2,
\end{equation}
where $y'$ is the output of feature-based decoder denoted by the blue shading in Figure \ref{fig:fig1}.

By optimizing the reconstruction loss in this hybrid manner, we obtain a meaningful low-dimensional latent space that captures the sequence structure relevant to the desired protein function. Additionally, our method comes with interpretability benefits that classical VAEs often lack. Existing works usually concatenate all features together into a single vector for the encoder. The resulting latent space is then obscured, as no physical meanings can be derived for the principal directions. In contrast, our hybrid training leads to meaningful visualizations of the data because the latent variables are directly linked to the chemical features.
 
Both the encoder and the decoder were implemented with multilayer perceptrons using PyTorch. Each has three fully connected layers with 256, 128, and 64 hidden units, respectively. The feature decoder has two fully connected layers with 32 hidden units. ReLU activation functions were used as non-linearities throughout the network, except in the output layer of the decoder where Sigmoid activation was used instead. The model was trained using the ADAM optimizer with a learning rate of 0.0001. A learning rate scheduler was used when validation loss stopped improving. 

\section{Results and Discussion}

\begin{figure*}[t]
    \centering
    \includegraphics[width=1\textwidth]{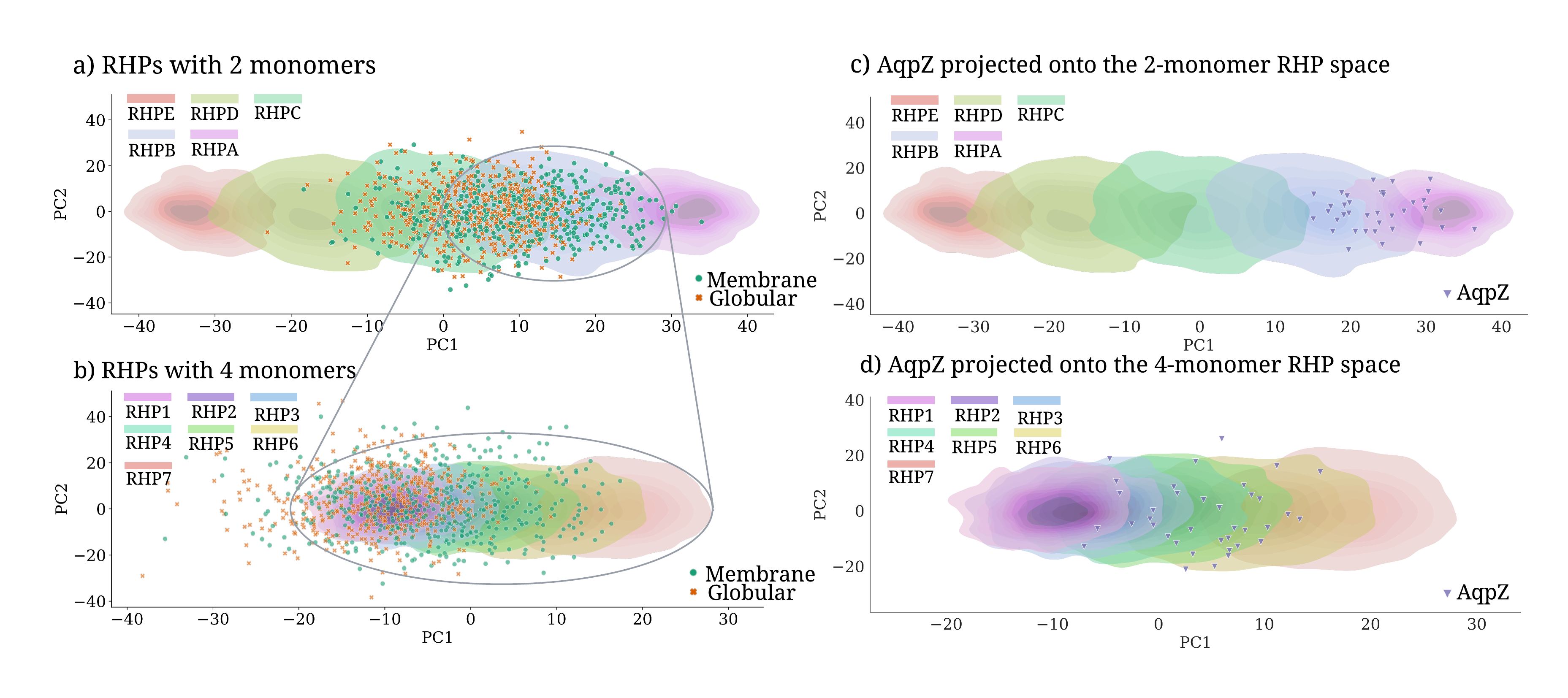}
    \caption{PCA projections of RHP and protein latent factors. Panels (a) and (b) project membrane and globular proteins onto two and four-monomer RHP space, respectively. Panels (c) and (d) project AqpZ onto the same two RHP spaces.}
    \label{fig:fig2}
\end{figure*}

Aquaporins (Aqp) are membrane channel proteins that facilitate water transport between cells. Membrane proteins are unstable and prone to aggregation even under mild experimental conditions. \citeauthor{Panganiban2018} successfully stabilized Aquaporin Z (AqpZ) and preserved its function in non-native environments with the presence of RHPs. We demonstrate how DeepRHP can be used to accelerate RHP design by identifying promising monomer compositions.

\citeauthor{Panganiban2018} \shortcite{Panganiban2018} chose to use 70\% hydrophobic monomers and 30\% hydrophilic monomers in their RHP system based on a crude protein surface analysis on four protein sequences. We first validate this distribution of monomer hydrophobicities using our model. The latent factors of the two-monomer RHPs and natural proteins are projected onto a two-dimensional space using Principal Component Analysis (PCA), as shown in Figure \ref{fig:fig2}(a). All two-monomer RHPs are composed of one hydrophobic monomer (EHMA) and one hydrophilic monomer (OEGMA). The compositions of \textit{RHP A} through \textit{RHP E} listed in Table \ref{table_comp} are selected to sufficiently reflect this hydrophobicity range. We observe that PC1 correlates with hydrophobicity as RHPs span left to right, with left being least hydrophobic to right being most hydrophobic. The majority of membrane and globular proteins overlap with \textit{RHP B} and \textit{RHP C}, suggesting these two RHP compositions are most similar to natural proteins. On the other hand, most hydrophobic membrane proteins overlap with \textit{RHP B} (30\% hydrophilic, 70\% hydrophobic), confirming that 30:70 is a good balance for the two-monomer system.

We then fine-tune the performance of the 30:70 distribution of hydrophilic and hydrophobic monomers by increasing the number of monomers from two to four as shown in Figure \ref{fig:fig2}(b). A library of four-monomer-based RHPs was designed by varying the MMA:EHMA ratio. The specific monomer composition is shown in Table \ref{table_comp}. Each of \textit{RHP 1} through \textit{RHP 7} is still composed of 30\% hydrophilic  monomers (OEGMA + SPMA) and 70\% hydrophobic monomers (MMA + EHMA).

\citeauthor{Panganiban2018} \shortcite{Panganiban2018} did not rationalize the choice of four monomers for their design of protein-like RHPs. Our approach explains why the two-monomer alphabet size is insufficient. In Figure \ref{fig:fig2}(b), each of the RHP ensembles can be considered as a subset of \textit{RHP B} and occupies a much more localized natural protein sequence space with smaller variance. In Figures \ref{fig:fig2}(c) and (d), we project AqpZ onto the two-monomer and four-monomer PCA spaces, respectively. In the two-monomer setting, the \textit{RHP B} space is much larger than the span of AqpZ. In the four-monomer setting, however, the AqpZ projections cover the \textit{RHP 4} and \textit{RHP 5} spaces almost entirely. Therefore, we believe the two-monomer sequence space is too broad with respect to proteins while the four-monomer sequence space is more localized, offering stability in synthesizing RHPs.

In addition to providing heuristics regarding the number of monomers, DeepRHP sheds light on the choice of monomer compositions. In Figure \ref{fig:fig2}(d), there is a large overlap between the projected proteins and the \textit{RHP 4} and \textit{RHP 5} contours. Wet-lab experiments in \citeauthor{Panganiban2018} \shortcite{Panganiban2018} demonstrated that the optimal RHP has the same monomer ratio as that of \textit{RHP 4} and is capable of stabilizing AqpZ. Thus, the overlap between RHPs and AqpZ in the PCA space can modulate their sequence correlation and molecular interactions in the aqueous solution. This indicates that the latent embeddings discovered by DeepRHP are chemically meaningful and play a key role in discovering RHPs that provide strong performance. 

\section{Conclusion}

In this study, we developed DeepRHP, a hybrid variational autoencoder model to guide RHP design. 
Our model suggests the feasibility of four-monomer compositions to stabilize ApqZ, matching the respective wet-lab experiment. In ablation studies, our model outperforms a singular classical VAE without the additional decoding regressor. 

Overall, DeepRHP holds much promise for the future of integrating deep learning techniques, specifically VAEs, into RHP design. Hybrid VAE architectures like DeepRHP possess many advantages. First, they are flexible and can be trained on any sequence family with variable sequence lengths and no multiple sequence alignment is needed. DeepRHP is also flexible due to its flexibility in supervision. It can be totally unsupervised when no prior knowledge on RHP subpopulations is available, or it can also be semi-supervised by combining function-related chemical features with vast amounts of sequence data to improve interpretability of latent variables. 

Future work in this regime includes strengthening the quantitative assessment of DeepRHP. Our model is currently assessed in a qualitative manner and validated using laboratory results. We hope to improve DeepRHP by developing a quantitative measure to evaluate the quality of the latent representations. For instance, we hope to complete further downstream tasks such as classifying specific membrane proteins and evaluating similarities between each RHP and their target proteins.

\section{Acknowledgments}
This work is supported by the U.S. Department of Defense (DOD), Army Research Office, under contract W911NF-13-1-0232 and the National Science Foundation under grant number DGE 2146752. We thank Yaodong Yu and Peter Bickel for useful discussions and comments on the model formulation.

\bibliography{aaai23.bib} 

@article{Hilburg2020,
   author = {Shayna L. Hilburg and Zhiyuan Ruan and Ting Xu and Alfredo Alexander-Katz},
   doi = {10.1021/ACS.MACROMOL.0C01886/SUPPL_FILE/MA0C01886_SI_001.PDF},
   issn = {15205835},
   issue = {21},
   journal = {Macromolecules},
   month = {11},
   pages = {9187-9199},
   publisher = {American Chemical Society},
   title = {Behavior of Protein-Inspired Synthetic Random Heteropolymers},
   volume = {53},
   url = {https://pubs.acs.org/doi/abs/10.1021/acs.macromol.0c01886},
   year = {2020},
}

@article{Panganiban2018,
   author = {Brian Panganiban and Baofu Qiao and Tao Jiang and Christopher DelRe and Mona M. Obadia and Trung Dac Nguyen and Anton A.A. Smith and Aaron Hall and Izaac Sit and Marquise G. Crosby and Patrick B. Dennis and Eric Drockenmuller and Monica Olvera De La Cruz and Ting Xu},
   doi = {10.1126/SCIENCE.AAO0335/SUPPL_FILE/AAO0335S2B.MP4},
   issn = {10959203},
   issue = {6381},
   journal = {Science},
   month = {3},
   pages = {1239-1243},
   pmid = {29590071},
   publisher = {American Association for the Advancement of Science},
   title = {Random heteropolymers preserve protein function in foreign environments},
   volume = {359},
   url = {https://www.science.org/doi/10.1126/science.aao0335},
   year = {2018},
}

@article{Jiang2020,
   author = {Tao Jiang and Aaron Hall and Marco Eres and Zahra Hemmatian and Baofu Qiao and Yun Zhou and Zhiyuan Ruan and Andrew D. Couse and William T. Heller and Haiyan Huang and Monica Olvera de la Cruz and Marco Rolandi and Ting Xu},
   doi = {10.1038/s41586-019-1881-0},
   issn = {1476-4687},
   issue = {7789},
   journal = {Nature},
   month = {1},
   pages = {216-220},
   pmid = {31915399},
   publisher = {Nature Publishing Group},
   title = {Single-chain heteropolymers transport protons selectively and rapidly},
   volume = {577},
   url = {https://www.nature.com/articles/s41586-019-1881-0},
   year = {2020},
}

@misc{Zhou2022,
  doi = {10.48550/ARXIV.2207.01813},
  url = {https://arxiv.org/abs/2207.01813},
  author = {Zhou, Yun and Gong, Boying and Jiang, Tao and Xu, Ting and Huang, Haiyan},
  title = {Stochastic Variational Methods in Generalized Hidden Semi-Markov Models to Characterize Functionality in Random Heteropolymers},
  publisher = {arXiv},
  year = {2022},
  copyright = {arXiv.org perpetual, non-exclusive license}
}

@misc{kingma_auto-encoding_2014,
  doi = {10.48550/ARXIV.1312.6114},
  url = {https://arxiv.org/abs/1312.6114},
  author = {Kingma, Diederik P and Welling, Max},
  title = {Auto-Encoding Variational Bayes},
  publisher = {arXiv},
  year = {2013},
  copyright = {arXiv.org perpetual, non-exclusive license}
}

@misc{rezende_stochastic_2014,
  doi = {10.48550/ARXIV.1401.4082},
  url = {https://arxiv.org/abs/1401.4082},
  author = {Rezende, Danilo Jimenez and Mohamed, Shakir and Wierstra, Daan},
  title = {Stochastic Backpropagation and Approximate Inference in Deep Generative Models},
  publisher = {arXiv},
  year = {2014},
  copyright = {arXiv.org perpetual, non-exclusive license}
}

@misc{Sinai2017,
  doi = {10.48550/ARXIV.1712.03346},
  url = {https://arxiv.org/abs/1712.03346},
  author = {Sinai, Sam and Kelsic, Eric and Church, George M. and Nowak, Martin A.},
  title = {Variational auto-encoding of protein sequences},
  publisher = {arXiv},
  year = {2017},
  copyright = {arXiv.org perpetual, non-exclusive license}
}

@article{Riesselman2018,
   author = {Adam J. Riesselman and John B. Ingraham and Debora S. Marks},
   doi = {10.1038/s41592-018-0138-4},
   isbn = {4159201801384},
   issn = {1548-7105},
   issue = {10},
   journal = {Nature Methods},
   month = {9},
   pages = {816-822},
   pmid = {30250057},
   publisher = {Nature Publishing Group},
   title = {Deep generative models of genetic variation capture the effects of mutations},
   volume = {15},
   url = {https://www.nature.com/articles/s41592-018-0138-4},
   year = {2018},
}

@misc{Costello2019,
  doi = {10.48550/ARXIV.1903.00458},
  
  url = {https://arxiv.org/abs/1903.00458},
  
  author = {Costello, Zak and Martin, Hector Garcia},
  
  title = {How to Hallucinate Functional Proteins},
  
  publisher = {arXiv},
  
  year = {2019},
  
  copyright = {arXiv.org perpetual, non-exclusive license}
}

@article{Greener2018,
   author = {Joe G. Greener and Lewis Moffat and David T. Jones},
   doi = {10.1038/s41598-018-34533-1},
   issn = {2045-2322},
   issue = {1},
   journal = {Scientific Reports},
   month = {11},
   pages = {16189},
   pmid = {30385875},
   publisher = {Nature Publishing Group},
   title = {Design of metalloproteins and novel protein folds using variational autoencoders},
   volume = {8},
   url = {https://www.nature.com/articles/s41598-018-34533-1},
   year = {2018},
}

@article{Smith2019,
   author = {Anton A.A. Smith and Aaron Hall and Vincent Wu and Ting Xu},
   doi = {10.1021/ACSMACROLETT.8B00813/SUPPL_FILE/MZ8B00813_SI_001.PDF},
   issn = {21611653},
   issue = {1},
   journal = {ACS Macro Letters},
   month = {1},
   pages = {36-40},
   pmid = {35619408},
   publisher = {American Chemical Society},
   title = {Practical Prediction of Heteropolymer Composition and Drift},
   volume = {8},
   url = {https://pubs.acs.org/doi/abs/10.1021/acsmacrolett.8b00813},
   year = {2019},
}

@article{uniprot,
    author = {UniProt Consortium, The},
    title = "{UniProt: the universal protein knowledgebase in 2021}",
    journal = {Nucleic Acids Research},
    volume = {49},
    Issue = {D1},
    pages = {D480-D489},
    year = {2020},
    month = {11},
    issn = {0305-1048},
    doi = {10.1093/nar/gkaa1100},
    url = {https://doi.org/10.1093/nar/gkaa1100},
    eprint = {https://academic.oup.com/nar/article-pdf/49/D1/D480/35364103/gkaa1100.pdf},
}

@article{DelRe2021,
   author = {Christopher DelRe and Yufeng Jiang and Philjun Kang and Junpyo Kwon and Aaron Hall and Ivan Jayapurna and Zhiyuan Ruan and Le Ma and Kyle Zolkin and Tim Li and Corinne D. Scown and Robert O. Ritchie and Thomas P. Russell and Ting Xu},
   doi = {10.1038/s41586-021-03408-3},
   issn = {1476-4687},
   issue = {7855},
   journal = {Nature},
   month = {4},
   pages = {558-563},
   pmid = {33883730},
   publisher = {Nature Publishing Group},
   title = {Near-complete depolymerization of polyesters with nano-dispersed enzymes},
   volume = {592},
   url = {https://www.nature.com/articles/s41586-021-03408-3},
   year = {2021},
}

@article{Tamasi2022,
   author = {Matthew J Tamasi and Roshan A Patel and Carlos H Borca and Shashank Kosuri and Heloise Mugnier and Rahul Upadhya and N Sanjeeva Murthy and Michael A Webb and Adam J Gormley and M J Tamasi and S Kosuri and H Mugnier and R Upadhya and N S Murthy and A J Gormley and R A Patel and C H Borca and M A Webb},
   doi = {10.1002/ADMA.202201809},
   issn = {1521-4095},
   issue = {30},
   journal = {Advanced Materials},
   month = {7},
   pages = {2201809},
   pmid = {35593444},
   publisher = {John Wiley & Sons, Ltd},
   title = {Machine Learning on a Robotic Platform for the Design of Polymer–Protein Hybrids},
   volume = {34},
   url = {https://onlinelibrary.wiley.com/doi/full/10.1002/adma.202201809 https://onlinelibrary.wiley.com/doi/abs/10.1002/adma.202201809 https://onlinelibrary.wiley.com/doi/10.1002/adma.202201809},
   year = {2022},
}

@article{Liang2022,
   author = {Yuchao Liang and Siqi Yang and Lei Zheng and Hao Wang and Jian Zhou and Shenghui Huang and Lei Yang and Yongchun Zuo},
   doi = {10.1016/J.CSBJ.2022.07.001},
   issn = {2001-0370},
   journal = {Computational and Structural Biotechnology Journal},
   month = {1},
   pages = {3503-3510},
   publisher = {Elsevier},
   title = {Research progress of reduced amino acid alphabets in protein analysis and prediction},
   volume = {20},
   year = {2022},
}

@article{Arnold2018,
   author = {Frances H. Arnold},
   doi = {10.1002/anie.201708408},
   issn = {15213773},
   issue = {16},
   journal = {Angewandte Chemie - International Edition},
   title = {Directed Evolution: Bringing New Chemistry to Life},
   volume = {57},
   year = {2018},
   pages = {4143-4148}
}

@article{Huang2016,
   author = {Po Ssu Huang and Scott E. Boyken and David Baker},
   doi = {10.1038/nature19946},
   issn = {1476-4687},
   issue = {7620},
   journal = {Nature},
   month = {9},
   pages = {320-327},
   pmid = {27629638},
   publisher = {Nature Publishing Group},
   title = {The coming of age of de novo protein design},
   volume = {537},
   url = {https://www.nature.com/articles/nature19946},
   year = {2016},
}

@article{Kyte1982,
   author = {Jack Kyte and Russell F. Doolittle},
   doi = {10.1016/0022-2836(82)90515-0},
   issn = {00222836},
   issue = {1},
   journal = {Journal of Molecular Biology},
   title = {A simple method for displaying the hydropathic character of a protein},
   volume = {157},
   year = {1982},
   pages = {105-132}
}

\end{document}